\documentclass[letterpaper]{article}
\usepackage{aaai19}
\usepackage{times}
\usepackage{helvet}
\usepackage{courier}
\usepackage{indentfirst}
\usepackage{natbib}
\usepackage{amssymb}
\usepackage{amsmath}
\usepackage{graphicx}
\usepackage{epstopdf}
\usepackage{array}
\usepackage{multirow}
\usepackage{float}
\usepackage{subfigure}
\usepackage{times}
\usepackage{bm}
\usepackage{tabularx}
\begin{document}
\title{Point2Sequence: Learning the Shape Representation of 3D Point Clouds with an Attention-based Sequence to Sequence Network}
\author{
Xinhai Liu\textsuperscript{1}, Zhizhong Han\textsuperscript{1,2}, Yu-Shen Liu\textsuperscript{1}\thanks{Corresponding author: Yu-Shen Liu}, Matthias Zwicker\textsuperscript{2}\\
\textsuperscript{1}School of Software, Tsinghua University, Beijing 100084, China \\
Beijing National Research Center for Information Science and Technology (BNRist)\\
\textsuperscript{2}Department of Computer Science, University of Maryland, College Park, USA\\
lxh17@mails.tsinghua.edu.cn, 
h312h@umd.edu, 
liuyushen@tsinghua.edu.cn, 
zwicker@cs.umd.edu
}
\maketitle
\begin{abstract}
Exploring contextual information in the local region is important for shape understanding and analysis.
Existing studies often employ hand-crafted or explicit ways to encode contextual information of local regions. However, it is hard to capture fine-grained contextual information in hand-crafted or explicit manners, such as the correlation between different areas in a local region, which limits the discriminative ability of learned features. To resolve this issue, we propose a novel deep learning model for 3D point clouds, named Point2Sequence, to learn 3D shape features by capturing fine-grained contextual information in a novel implicit way. Point2Sequence employs a novel sequence learning model for point clouds to capture the correlations by aggregating multi-scale areas of each local region with attention. Specifically, Point2Sequence first learns the feature of each area scale in a local region. Then, it captures the correlation between area scales in the process of aggregating all area scales using a recurrent neural network (RNN) based encoder-decoder structure, where an attention mechanism is proposed to highlight the importance of different area scales. Experimental results show that Point2Sequence achieves state-of-the-art performance in shape classification and segmentation tasks.

\end{abstract}

\section{Introduction}
3D point clouds, also called point sets, are considered as one of the simplest 3D shape representations, since they are composed of only raw coordinates in 3D space.
A point cloud can be acquired expediently by popular sensors such as LiDAR, conventional cameras, or RGB-D cameras. 
Furthermore, this kind of 3D data is widely used in 3D modeling (\citealt{golovinskiy2009shape}), autonomous driving (\citealt{qi2017frustum}), indoor navigation (\citealt{zhu2017target}) and robotics (\citealt{wang2015voting}).
However, learning features or shape representations based on point clouds by deep learning models remains a challenging problem due to the irregular nature of point clouds (\citealt{qi2017pointnet}).

As a pioneering approach, PointNet (\citealt{qi2017pointnet}) resolves this challenge by directly applying deep learning on point sets.
PointNet individually computes a feature of each point and then aggregates all the point features into a global feature by a pooling layer.
This leads to PointNet being limited by capturing contextual information of local regions.
Attempting to address this issue, several researches take the aggregation of local regions into consideration.
KC-Net (\citealt{shen2018mining}) employs a kernel correlation layer and a graph-based pooling layer to capture the local information of point clouds.
SO-Net (\citealt{li2018so}) and DGCNN (\citealt{wang2018dynamic}) further explore the local structures by building k-nearest neighbors (kNN) graphs and integrating neighbors of a given point using learnable edge attributes in the graph.
PointNet++ (\citealt{qi2017pointnet++}) first extracts features for multi-scale local regions individually and aggregates these features by concatenation, where the two steps are repeated to complete the hierarchical feature extraction.
Similarly, ShapeContextNet (\citealt{xie2018attentional}) segments the local region of a given point into small bins, then extracts the feature for each bin individually, and finally concatenates the features of all bins as the updated feature for the point.
However, most of these previous methods employ hand-crafted or explicit ways for encoding contextual information in local regions, which makes it hard to fully capture fine-grained contextual information, such as the correlation between different areas in the feature space. 
However, the correlation between different areas in a local region is an important contextual information. Fully exploiting this information might enhance the discriminability of learned features and improve the performance in shape analysis tasks.

We address this issue by proposing a novel deep learning model for 3D point clouds, called Point2Sequence, to encode fine-grained contextual information in local regions in a novel implicit way.
Point2Sequence employs a novel RNN-based sequence model for local regions in point clouds to capture the correlation by aggregating multi-scale areas with attention.
Specifically, each local region is first separated into multi-scale areas. 
Then, the feature of each area scale is extracted by a shared Multi-Layer-Perceptron (MLP) layer.
Finally, our novel encoder-decoder based sequence model aggregates the features of all area scales, where an attention mechanism is involved to highlight the importance of different scale areas.  
Experimental results show that Point2Sequence is able to learn more discriminative features from point clouds than existing methods in shape classification and part segmentation tasks.


Our contributions are summarized as follows.
\begin{itemize}
\item 
We propose Point2Sequence to learn features from point clouds by capturing correlations between different areas in a local region, which takes full advantage of encoding fine-grained contextual information in local regions. 
\item We introduce an attention mechanism to highlight the importance of different scale areas, and the feature extraction of local regions is enhanced by leveraging the correlation between different area scales.  
\item To the best of our knowledge, Point2Sequence is the first RNN-based model for capturing correlations between different areas in local regions of point clouds, and our outperforming results verify the feasibility of RNNs to effectively understand point clouds.
\end{itemize}

\section{Related Work}
\noindent
\textbf{Learning from point clouds by rasterization.}\quad  As an irregular type of 3D data, it is intuitive to rasterize the point clouds into uniform sparse 3D grids and then apply volumetric convolutional neural networks.
Some approaches (\citealt{wu20153d,maturana2015voxnet}) represent each voxel with a binary representation which indicates whether it is occupied in the space.
Nevertheless, the performance is largely limited by the time and memory consuming due to the data sparsity of 3D shapes.
Several improvements (\citealt{li2016fpnn,wang2015voting}) have been proposed to relief the sparsity problem of the volumetric representation.
However, the sparsity of 3D shapes is an inherent drawback, which still makes it difficult to process very large point clouds.
Some recent methods (\citealt{su2015multi,wang2017dominant}) have tried to project the 3D point clouds or 3D shapes into 2D views and then apply 2D CNNs to recognize them.
Influenced by the great success of 2D CNNs for images, such methods have achieved dominant results in 3D shape classification and retrieval tasks (\citealt{kanezaki2016rotationnet}).
Due to the lack of depth information, it is nontrivial to extend view-based methods to per-point processing tasks such as point classification and shape classification.

Compared with uniform 3D grids, some latest studies utilize more scalable indexing techniques such as kd-tree and octree to generate  regular structures which can facilitate the use of deep learning functions.
To enable 3D convolutional networks, OctNet (\citealt{riegler2017octnet}) build a hierarchically partition of the space by generating a set of unbalanced octrees in the regular grids, where each leaf node stores a pooled feature representation.
Kd-Net (\citealt{klokov2017escape}) performs multiplicative transformations according to the subdivisions of point clouds based on the kd-trees.
However, in order to obtain rotation invariant of shapes,  these methods usually require extra operations such as pre-alignment or excessive data argumentations.

Different from the above-mentioned methods, our method directly learns from point clouds without pre-alignment and voxelization.

\noindent
\newline
\textbf{Learning from point clouds directly.}\quad  As a pioneer, PointNet (\citealt{qi2017pointnet}) achieves satisfactory performance by directly applying deep learning methods on point sets.
PointNet individually computes the feature of each point and then aggregates all the point features into a global feature by pooling.
This leads to PointNet limited by capturing contextual information in local regions.
Some enhancements are proposed to address this problem by combining contextual information in local regions by hand-crafted or explicit ways.
PointNet++ (\citealt{qi2017pointnet++}) has been proposed to group points into several clusters in pyramid-like layers, where the feature of multi-scale local regions can be extracted hierarchically.
PointCNN (\citealt{li2018pointcnn}) and SpiderCNN (\citealt{xu2018spidercnn}) investigate the convolution-like operations which aggregate the neighbors of a given point by edge attributes in the local region graph.
However, with hand-crafted or explicit ways of encoding contextual information in local regions, it is hard for these methods to capture fine-grained contextual information.
In particular, the correlation between different areas in feature space is an important contextual information, which limits the discriminative ability of learned features for point cloud understanding.
\begin{figure}[tp]
\centering 
\includegraphics[width=0.8\linewidth]{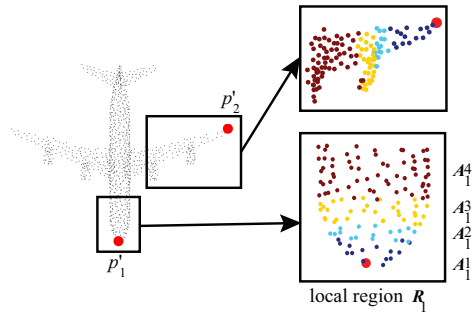}
\caption{The \textit{left} is an airplane point cloud with two sampled centroids $p'_1$ and $p'_2$ in the red color. The \textit{right} is the corresponding local regions of $p'_1$ (below) and $p'_2$ (above), where different colors represent different area scales within the local region. For example, there are four different scale areas $\lbrace \bm{A}_1^1, \bm{A}_1^2, \bm{A}_1^3, \bm{A}_1^4 \rbrace$ in the local region $\bm{R}_1$ centered by $p'_1$.}
\label{fig:multi-scalas}  
\end{figure}
\begin{figure*}[tp]
\centering
\includegraphics[height=8cm,width=18cm]{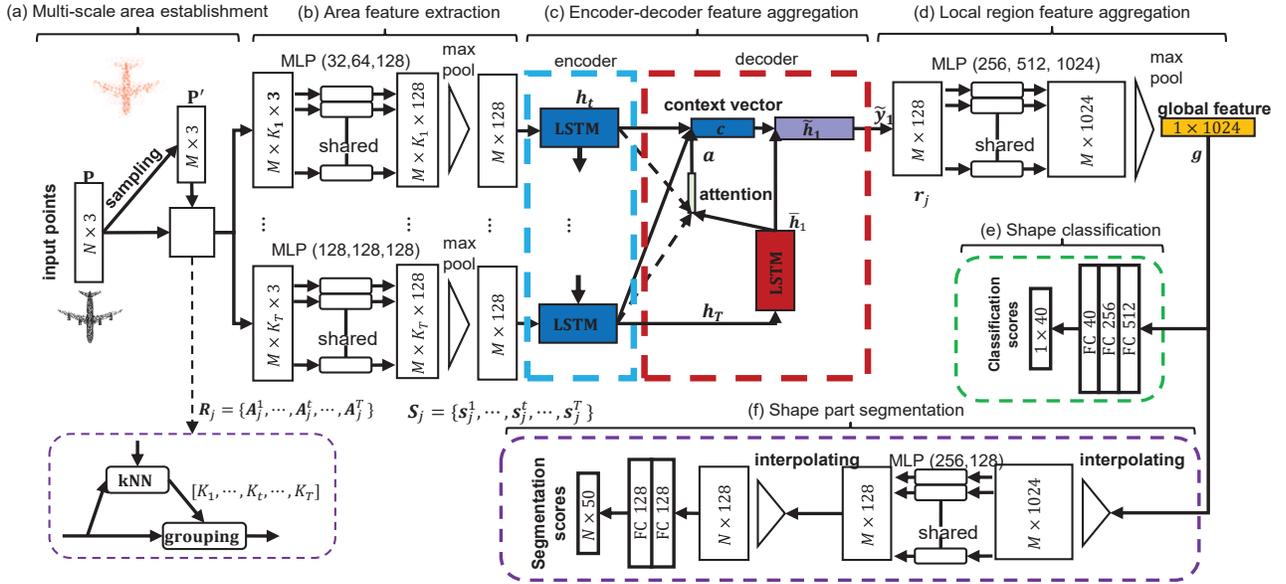}
\caption{\textbf{Our Point2Sequence architecture.}\quad Point2Sequence first samples local regions from an input point cloud and establishes multi-scale areas in each local region in (a). Then, MLP layer is employed to extract the feature of each multi-scale area in (b). Subsequently, the feature of each local region is extracted by attention-based seq2seq structure in (c). Finally, the global feature of the point cloud is obtained by aggregating the features of all sampled local regions in (d). The learned global feature can be used not only for shape classification shown in (e) but also for part segmentation with some extension network shown in (f).
}
\label{fig:network_arichitecture} 
\end{figure*}

\noindent
\newline
\textbf{Correlation learning with RNN-based models.}\quad To aggregate sequential feature vectors, recurrent neural networks (\citealt{elman1990finding}) have shown preeminent performance in many popular tasks such as speech recognition (\citealt{li2015constructing}) or  handwriting recognition (\citealt{bertolami2009novel}).
Inspired by the sequence to sequence architecture (\citealt{cho2014learning}), RNN-based seq2seq models can capture the fine-grained contextual information from the input sequence and effectively convert it to another sequence.
Furthermore, \citet{sutskever2014sequence} engages an attention mechanism that intuitively allows neural networks to focus on different parts of the input sequences, which is more in line with the actual situation.
In addition, several kinds of attention mechanism (\citealt{bahdanau2014neural,luong2015effective}) have been presented to enhance the performance of networks in machine translation (MT). 
To utilize the powerful ability of correlation learning from RNN-based sequence to sequence structure, Point2Sequence adopts a seq2seq encoder-decoder architecture to learn the correlation of different areas in each local region with attention. 

\section{The Point2Sequence Model}
In this section, we first overview our Point2Sequence model, and then detail the model including multi-scale area establishment, multi-scale area feature extraction, attention-based sequence to sequence structure and model adjustments for segmentation. 
\subsection{Overview}
Figure \ref{fig:network_arichitecture} illustrates our Point2Sequence architecture. Our model is  formed by six parts: (a) Multi-scale area establishment, (b) Area feature extraction, (c) Encoder-decoder feature aggregation, (d) Local region feature aggregation, (e) Shape classification and (f) Shape part segmentation.
The input of our network is a raw point set $\mathbf{P}=\lbrace p_i \in \mathbb{R}^3, i=1,2,\cdots,N\rbrace$.
We first select $M$ points 
$\mathbf{P}'=\lbrace p'_j \in \mathbb{R}^3, j=1,2,\cdots,M\rbrace$
from $\mathbf{P}$ to define the centroids of local regions $\lbrace \bm{R}_1,\cdots, \bm{R}_j, \cdots, \bm{R}_M \rbrace$.
As illustrated in Figure \ref{fig:multi-scalas}, $T$ different scale areas $\lbrace \bm{A}_j^1, \cdots, \bm{A}_j^t, \cdots, \bm{A}_j^T \rbrace$ in each local region $\bm{R}_j$ centered by $p'_j$ are established in the multi-scale area establishment part, where $T$ different scale areas contain $[K_1,\cdots, K_t, \cdots,K_T]$ points, respectively.
We then extract a $D$-dimensional feature $\bm{s}_j^t$ for each scale area $\bm{A}_j^t$  through the scale feature extraction part.
In each local region $\bm{R}_j$, a feature sequence $\bm{S}_j=\lbrace \bm{s}_j^1,\cdots, \bm{s}_j^t, \cdots, \bm{s}_j^T\rbrace$ of multi-scale areas is aggregated into a $D$-dimensional feature vector $\bm{r}_j$ by the encoder-decoder feature aggregation part.
Finally, a $1024$-dimensional global feature $\bm{g}$ is aggregated from the features $\bm{r}_j$ of all local regions by the local region feature aggregation part.
Our model can be trained for shape classification or shape part segmentation by minimizing the error according to the ground truth class labels or per point part labels.  

\subsection{Multi-scale Area Establishment}
Similar to PointNet++ (\citealt{qi2017pointnet++}) and ShapeContextNet (\citealt{xie2018attentional}), we build the structure of multi-scale areas in each local region on the point cloud. This part is formed by three layers: sampling layer, searching layer and grouping layer.
The sampling layer selects $M$ points from the input point cloud as the centroids of local regions. 
In each local region, searching layer searches the $K_t$ points in the input points and return the corresponding point indexes. 
According to the point indexes, grouping layer groups the $K_t$ points from $\mathbf{P}$ to form each multi-scale area.

As shown in Figure \ref{fig:network_arichitecture},  a certain amount of points are first selected as the centroids of local regions by the sampling layer.
We adopt the farthest point sampling (FPS) to iteratively select $M$ ($M < N$) points. 
The new added point $p'_j$ is always the farthest point from points $\lbrace p'_1, p'_2, \cdots, p'_{j-1} \rbrace$ in the 3D space.
Although random sampling is an optional choice, the FPS can achieves a more uniform coverage of the entire point cloud in the case of the same centroids.

Then the searching layer respectively finds the top $[K_1,\cdots, K_t, \cdots,K_T]$ nearest neighbors for each local region $\bm{R}_j$ from the input points and returns the corresponding indexes of these points.
In the searching layer, we adopt kNN to find the neighbors of centroids according to the sorted Euclidean distance in the 3D space. 
Another optional search method is ball search (\citealt{qi2017pointnet++}) which selects  all points within a radius around the centroid.
Compared with ball search, kNN search guarantees the size of local regions and makes it insensitive to the sampling density of point clouds.

Finally, by the grouping layer, these indexes of points are used to extract the points in each area of local region $\bm{R}_j$, where we obtain the points with the size of $M \times K_t \times 3$ in the scale area $\bm{A}_j^t$ of all local regions.

\subsection{Multi-scale Area Feature Extraction}
To extract the feature for each multi-scale area, a simple and effective MLP layer is employed in our network.
The MLP layer first abstracts the points in each scale area $\bm{A}_j^t$ into the feature space and then the point features are aggregated into a $D$-dimensional feature $\bm{s}_j^t$ by max pooling.
Therefore, in each local region $\bm{R}_j$, a $T \times D$ feature sequence $\bm{S}_j$ of the multi-scale areas is acquired.

In addition, similar to prior studies such as SO-Net (\citealt{li2018so}), the coordinates of points in each area $\bm{A}_j^t$ are converted to the relative coordinate system of the centroid $p'_j$ by a subtraction operation: $p_{l} = p_{l} - p'_j$, where $l$ is the index of point in the area.
By using the relative coordinate, the learned features of point sets can be invariant to transformations such as rotation and translation, which is crucial for improving the learning ability of networks.
Moreover, we also combine the area feature $\bm{s}_j^t$ with the coordinates of area centroid $p'_j$ to enhance the association between them.

\subsection{Attention-based Sequence to Sequence Structure}
In this subsection, we propose an attention-based encoder-decoder network to capture the fine-grained contextual information of local regions.
Through the multi-scale area feature extraction, each local region $\bm{R}_j$ is abstracted into a feature sequence $\bm{r}_j$. 
To learn from the feature sequence, we adopt a RNN-based model to aggregate the features $\bm{S}_j$ of size $T \times D$ into a $D$ dimension feature vector $\bm{s}_j$ in each local region by an implicit way. 
To further promote the correlation learning between different areas in local regions, we employ an attention mechanism to focus on items of the feature sequence.
\begin{table*}[htp]
\centering
\caption{The shape classification accuracy (\%) comparison on ModelNet10 and ModelNet40.}
\label{table:compare}
\begin{tabular}{l|c|cc|cc}
\hline 
\multirow{2}{*}{Method}&
    \multicolumn{1}{c|}{\multirow{2}{*}{Input}} & 
	\multicolumn{2}{c|}{ModelNet10} &
	\multicolumn{2}{c}{ModelNet40} \\ 
    & \multicolumn{1}{c|}{}
    & \multicolumn{1}{c}{Class}
    & \multicolumn{1}{c|}{Instance}
    & \multicolumn{1}{c}{Class}
    & \multicolumn{1}{c}{Instance} \\ \hline
 PointNet (\citealt{qi2017pointnet})    				          &$1024 \times 3  $ &-     &-    &86.2 &89.2 \\ 
 PointNet++ (\citealt{qi2017pointnet++})				          &$1024 \times 3  $ &-     &-    &-    &90.7 \\ 
 ShapeContextNet (\citealt{xie2018attentional})					  &$1024 \times 3  $ &-		&-	  &87.6	&90.0 \\
 Kd-Net (\citealt{klokov2017escape})    				          &$2^{15} \times 3$ &93.5  &94.0 &88.5 &91.8 \\ 
 KC-Net (\citealt{shen2018mining})  				          &$1024 \times 3$   &-     &94.4 &-    &91.0 \\ 
 PointCNN (\citealt{li2018pointcnn})    				        &$1024 \times 3$   &-     &-    &-    &91.7 \\ 
 DGCNN (\citealt{wang2018dynamic})      			         &$1024 \times 3$   &-     &-    &90.2 &92.2 \\
 SO-Net (\citealt{li2018so})            				          &$2048 \times 3$   &93.9  &94.1 &87.3 &90.9 \\ \hline
 Ours                               				          &$1024 \times 3$   &\textbf{95.1}  &\textbf{95.3} &\textbf{90.4} &\textbf{92.6} \\
\hline\end{tabular}
\end{table*}

\noindent
\newline
\textbf{Multi-scale area encoder.}\quad To learn the correlation between different areas, a RNN is employed as encoder to integrate the features of multi-scale areas in a local region. The RNN encoder is consisted by a hidden state $\mathbf{h}$ and an optional output $\mathbf{y}$, which operates on the input feature sequence $\bm{S}_j = \lbrace \bm{s}_j^1, \dots, \bm{s}_j^t, \dots, \bm{s}_j^T \rbrace$ of multi-scale areas in each local region. Each item $\bm{s}_j^t$ of the sequence $\bm{S}_j$ is a $D$-dimensional feature vector and the length of $\bm{S}_j$ is $T$ which is also the steps of the encoder.  
At each time step $t$, the hidden state $\mathbf{h}_{t}$ of the RNN encoder is updated by
\begin{equation}
\mathbf{h}_{t} = f(\mathbf{h}_{t-1}, \bm{s}_j^t),
\label{equal:eh}
\end{equation}
where $f$ is a non-linear activation function which can be a long short-term memory (LSTM) unit (\citealt{hochreiter1997long}) or a gated recurrent unit (\citealt{cho2014learning}).

A RNN can learn the probability distribution over a sequence by being trained to predict the next item in the sequence.
Similarly, at time $t$, the output $\mathbf{y}_t$ of the encoder can be represented as
\begin{equation}
\mathbf{y}_{t} = \mathbf{W}_a\mathbf{h}_t,
\label{equal:ey}
\end{equation}
where $\mathbf{W}_a$ is a learnable weight matrix. 

After forwarding the entire input feature sequence at step $T$, the last-step  hidden state $\mathbf{h}_T$ of the encoder is acquired, which contains the context information of the entire input sequence.

\begin{table*}
\caption{The accuracies (\%) of part segmentation on ShapeNet part segmentation dataset.}
\label{table:part_segmentaion}
\resizebox{\textwidth}{19mm}{
\begin{tabular}{l|c|cccccccccccccccccccc}
\hline 
\multirow{2}{*}{}&
	\multicolumn{1}{c|}{\multirow{2}{*}{mean}} & 
	\multicolumn{16}{c}{Intersection over Union (IoU)}\\ 
	& \multicolumn{1}{c|}{}
    & \multicolumn{1}{c}{air.}
    & \multicolumn{1}{c}{bag}
    & \multicolumn{1}{c}{cap}
    & \multicolumn{1}{c}{car}
    & \multicolumn{1}{c}{cha.}
    & \multicolumn{1}{c}{ear.}
    & \multicolumn{1}{c}{gui.}
    & \multicolumn{1}{c}{kni.}
    & \multicolumn{1}{c}{lam.}
    & \multicolumn{1}{c}{lap.}
    & \multicolumn{1}{c}{mot.}
    & \multicolumn{1}{c}{mug}
    & \multicolumn{1}{c}{pis.}
    & \multicolumn{1}{c}{roc.}
    & \multicolumn{1}{c}{ska.}
    & \multicolumn{1}{c}{tab.}
    \\ \hline
\# SHAPES &   &2690 &76 &55 &898 &3758 &69 &787 &392 &1547 &451 &202 &184 &283 &66 &152 &5271 \\ \hline
PointNet 			&83.7&83.4&78.7&82.5&74.9&89.6&73.0&91.5&85.9&80.8&95.3&65.2&93.0&81.2&57.9&72.8&80.6 \\ \hline
PointNet++  		&85.1&82.4&79.0&87.7&77.3&90.8&71.8&91.0&85.9&83.7&95.3&\textbf{71.6}&94.1&81.3&58.7&\textbf{76.4}&82.6 \\ \hline
ShapeContextNet &84.6 &83.8 &80.8 &83.5 &\textbf{79.3} &90.5 &69.8 &\textbf{91.7} &86.5 &82.9 &\textbf{96.0 }&69.2 &93.8 &82.5 &\textbf{62.9} &74.4 &80.8 \\ \hline
Kd-Net				&82.3&80.1&74.6&74.3&70.3&88.6&73.5&90.2&87.2&81.0&94.9&57.4&86.7&78.1&51.8&69.9&80.3 \\ \hline
KCNet   			&84.7&82.8&81.5&86.4&77.6&90.3&76.8&91.0&87.2&\textbf{84.5}&95.5&69.2&94.4&81.6&60.1&75.2&81.3 \\ \hline

DGCNN &85.1 &\textbf{84.2} &\textbf{83.7} &84.4 &77.1 &\textbf{90.9} &\textbf{78.5} &91.5 &\textbf{87.3} &82.9 &\textbf{96.0} &67.8 &93.3 &\textbf{82.6} &59.7 &75.5 &82.0	\\ \hline
SO-Net 				&84.9&82.8&77.8&\textbf{88.0}&77.3&90.6&73.5&90.7&83.9&82.8&94.8&69.1&94.2&80.9&53.1&72.9&\textbf{83.0} \\ \hline
Ours	&\textbf{85.2}&82.6&81.8&87.5&77.3&90.8&77.1&91.1&86.9&83.9&95.7&70.8&\textbf{94.6}&79.3&58.1&75.2&82.8 \\ \hline
\end{tabular}}
\end{table*}
\noindent
\newline
\textbf{Local region feature decoder.}\quad To obtain the feature $\bm{r}_j$ for each local region $\bm{R}_j$, we employ a RNN decoder to translate the contextual information from the multi-scale areas in $\bm{R}_j$. 
Different from the models in machine translation, there is no decoding target for the decoder in our case.
To address this issue, we employ $\mathbf{h}_T$ as the decoding target which contains contextual information of the entire input feature sequence.
Therefore, a one-step decoding process is adopt in our network to decode the feature $\bm{r}_j$ for each local region $\bm{R}_j$.
Similar to the encoder, the decoder is also consisted by a hidden state $\mathbf{\bar{h}}$ and output $\mathbf{\bar{y}}$.
We initialize the $\mathbf{\bar{h}}_0$ with a zero state $\mathbf{z}_0$ and the current hidden state of the decoder at step one can be updated by
\begin{equation}
\mathbf{\bar{h}}_1 = f(\mathbf{z}_0, \mathbf{h}_T),
\end{equation}
where $f$ is an activation function as shown in Eq. (\ref{equal:eh}).
Similarly, the output of decoder $\mathbf{\bar{y}}_1$ is computed by
\begin{equation}
\mathbf{\bar{y}}_1 = \mathbf{W}_b \mathbf{\bar{h}}_1,
\end{equation}
where $\mathbf{W}_b$ is a learnable matrix in the training.

To further enhance the decoding of the contextual information in the input feature sequence $\bm{S}_j$, a context vector $\mathbf{c}$ is generated to help the predict of feature $\mathbf{\tilde{y}}_1$ of local region with attention. Therefore, a new hidden state $\mathbf{\tilde{h}_1}$ and output $\mathbf{\tilde{y}}_1$ are computed in the decoder as introduced later, where we employ the output $\mathbf{\tilde{y}}_1$ to be the feature $\bm{r}_j$ for each local region. 


\noindent
\newline
\textbf{Attention mechanism in decoder.}\quad Inspired by the thought of focusing on parts of the source sentence in the machine translation, we adopt an attention mechanism to highlight the importance of different areas in each local region.
The goal is to utilize the context vector $\mathbf{c}$ which is generated by
\begin{equation}
\mathbf{c} = {\sum}^T_{t=1}{\bm{\alpha}(t)\mathbf{h}_t},
\label{equal:context}
\end{equation}
where $\bm{\alpha}$ is the attention vector and $t$ is the time step.

The idea of our attention model is to consider all the hidden states of the encoder when generating the context vector $\mathbf{c}$.
In this model, a fixed-length attention vector $\bm{\alpha}$, whose size is equal to the sequence length $T$ of the input side, is generated by comparing the current hidden state $\mathbf{\bar{h}_1}$ with each source hidden state $\mathbf{h}_t$ as
\begin{equation}
\begin{aligned}
\bm{\alpha}(t) &= \frac{exp(score(\mathbf{\bar{h}}_1, \mathbf{h}_t))}{{\sum}^T_{t^{'}=1}{exp(score(\mathbf{\bar{h}}_1, \mathbf{h}_{t^{'}}))}}.
\end{aligned}
\end{equation}
Here, $score$ is referred as 
\begin{equation}
score(\mathbf{\bar{h}}_1, \mathbf{h}_t) = {\mathbf{\bar{h}}_1}^{\top} \mathbf{W}_c \mathbf{h}_t,
\end{equation}
which is a content-based function to show the correlation between these two vectors. Here, $\mathbf{W}_c$ is also a learnable weight matrix in the training. 

With the help of Eq. (\ref{equal:context}), we can acquire the new hidden state $\mathbf{\tilde{h}}_1$ based on the current hidden state $\mathbf{\bar{h}}_1$ in the decoder.
Specifically, with the current hidden state $\mathbf{\bar{h}}_1$ and the context vector $\mathbf{c}$, a simple concatenation layer combines the  contextual information of the two vectors to generate the attentional new hidden state as follows,
\begin{equation}
\mathbf{\tilde{h}}_1 = tanh(\mathbf{W}_d[\mathbf{c}; \mathbf{\bar{h}}_1]).
\end{equation}
And the output $\mathbf{\tilde{y}}_1$ is similarly computed by
\begin{equation}
\mathbf{\tilde{y}}_1 = \mathbf{W}_s\mathbf{\tilde{h}}_1,
\end{equation}
where $\mathbf{W}_d$ and $\mathbf{W}_s$ are variables to be learned.

Our attention-based sequence to sequence structure aggregates the sequential features $\bm{S}_j$ of multi-scale areas in each local region $\bm{R}_j$ by an implicit way. So far, the features $\bm{r}_j$ of all local regions with size of $M \times D$ are acquired. 
In the subsequent network, a 1024-dimensional global feature $\bm{g}$ of the input point cloud is extracted from the features of all local regions by the global feature extraction part.
As depicted in Figure \ref{fig:network_arichitecture}, we apply the global feature $\bm{g}$ to shape classification and part segmentation tasks. 

\subsection{Model Adjustments for Segmentation}
The goal of part segmentation is to predict a semantic label for each point in the point cloud.
In Point2Sequence, a global feature of the input point cloud is generated.
Therefore, we need to acquire the per-point feature for each point in the point cloud from the global feature.
There are two optional implementations, one is to duplicate the global feature with $N$ times (\citealt{qi2017pointnet,wang2018dynamic}), and the other is to perform upsampling by interpolation (\citealt{qi2017pointnet++,li2018so}).
In this paper, two interpolate layers are equipped in our networks as shown in Figure \ref{fig:network_arichitecture}, which propagate the feature from shape level to point level by upsampling.
Compared with shape classification, it is a challenge task to distinguish the parts in the object, which requires more fine-grained contextual information of local regions. 
We implement the feature $\phi$ propagation according to the Euclidean distance between points in 3D space.
The feature is interpolated by inverse square Euclidean distance wighted average based on $k$ nearest neighbors as
\begin{equation}
\phi(p) = \frac{\sum_{i=1}^{k}{w(p_i)\phi(p_i)}}{\sum_{i=1}^{k}{w(p_i)}},
\label{equation:interpolate}
\end{equation}
where  $w(p_i)=\frac{1}{(p-p_i)^2}$ is the inverse square Euclidean distance between $p$ and $p_i$.

To guide the interpolation process, the interpolated features are concatenated with the corresponding point features in the abstraction side and several MLP layers are equipped in our network to enhance the performance.
Several shared fully connected layers and ReLU layers are also applied to promote the extraction of point features, like the branch in shape classification.

\begin{table}[tp]
\centering
\caption{The effect of the number of sampled points $M$ on ModelNet40.}
\label{table:sample_points}
\begin{tabular}{cccccc}\hline
$M$&128&256&384&512\\ \hline
Acc (\%)&91.86&92.34&\textbf{92.54}&91.86\\ \hline
\end{tabular}
\end{table}
\section{Experiments}
In this section, we first investigate how some key parameters affect the performance of Point2Sequence in the shape classification task.
Then, we compare our Point2Sequence with several state-of-the-art methods in shape classification on ModelNet10 and ModelNet40 (\citealt{wu20153d}), respectively.
Finally, the performance of our Point2Sequence is evaluated in the part segmentation task on ShapeNet part dataset (\citealt{savva2016shrec}).

\subsection{Ablation Study}
\noindent
\textbf{Network configuration.}\quad In Point2Sequence, we first sample $M = 384$ points as the centroids of local regions by the sampling layer.
Then the searching layer and grouping layer select $T=4$ scale of areas with $[16, 32, 64, 128]$ points in each area of a local region.
The points in each area are abstracted into a $D=128$ dimensional feature by a 3-layer MLP and then these abstracted features are aggregated by max pooling.
And the feature sequence of different areas in each local region is aggregated by the RNN-based encoder-decoder structure with attention.
Here, we initialize the RNN encoder and decoder with h=128 dimensional hidden state, where LSTM is used as the RNN cell.
The rest setting of  our network is the same as in Figure \ref{fig:network_arichitecture}.
In addition, ReLU is used after each fully connected layer with Batch-normalization, and Dropout is also applied with drop ratio 0.4 in the fully connected layers. 
In the experiment, we trained our network on a NVIDIA GTX 1080Ti GPU using ADAM optimizer with initial learning rate 0.001, batch size of 16 and batch normalization rate 0.5.
The learning rate and batch normalization rate are decreased by 0.3 and 0.5 for every 20 epochs, respectively.
\begin{figure*}[htp]
\centering 
\includegraphics[width=0.9\linewidth]{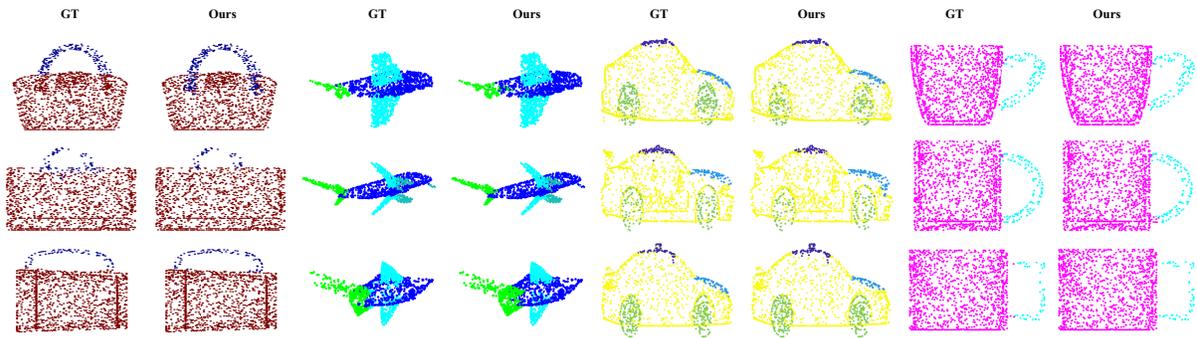}
\caption{Visualization of part segmentation results. In each shape pair, first column is the ground truth (GT), and second column is our predicted result, where parts with the same color have a consistent meaning. From left to right: bag, airplane, car, and cup.} 
\label{fig:part_segmentation}  
\end{figure*}

\noindent
\newline
\textbf{Parameters.}\quad All the experiments in the ablation study are evaluated on ModelNet40. ModelNet40 contains 12311 CAD models from 40 categories and is split into 9843 for training and 2468 for testing.
For each model, we adopt 1024 points which are uniformly sampled from mesh faces and are normalized into a unit ball as input.

We first explore the number of sampled points $M$ which influences the distribution of local regions in point clouds. 
In the experiment, we keep the settings of our network as depicted in the network configuration section and modifies the number of sampled points $M$ from 128 to 512.
The results are shown in Table\ref{table:sample_points}, where the instance accuracies on the benchmark of ModelNet40 have a tendency to rise first and then fall. 
The highest accuracy of $92.54\%$ is reached at $M=384$ sampled points.
This comparison implies that Point2Sequence can extract the contextual information from local regions effectively and $M=384$ is a optimum number of sampled points to coverage the whole point cloud.

\begin{table}[htp]
\centering
\caption{The effects of the type of RNN cell (CT) and hidden state dimension $h$ on ModelNet40.}
\label{table:rnn}
\begin{tabular}{cccccc}\hline
Metric&RT=LSTM&GRU&$h$=64&256\\ \hline
Acc (\%)&\textbf{92.54}&92.18&92.46&92.18\\ \hline
\end{tabular}
\end{table}
Therefore, we employ the sampled points $M = 384$ as the setting of our network in the following experiments. Then, as shown in Table \ref{table:rnn}, we show the effect of the type of the RNN cell (RT) and the dimension of the RNN hidden state $h$, respectively.
The accuracy degenerates to $92.18\%$ when replacing the LSTM cell as the GRU cell.
Based on the setting of $h = 128$, we set $h$ to 64 and 256, which reduce the accuracy of $h=128$ to $92.46\%$ and $92.18\%$.
The above results suggest that dimension of hidden state $h = 128$ and the type of hidden state RT=LSTM is more suitable for our network.

\begin{table}[htp]
\centering
\caption{The effects of the attention mechanism (Att) and decoder (Dec) on ModelNet40.}
\label{table:att}
\begin{tabular}{cccccc}\hline
Metric&Att+ED&No Att&No Dec &Con &MP \\ \hline
Acc (\%)&\textbf{92.54}&92.26&92.42&92.06 &91.73\\ \hline
\end{tabular}
\end{table}
We also discuss the impact of the attention mechanism, the decoder and the encoder-decoder structure on our network.
We evaluate the performance of our network without the attention mechanism (No Att), without decoder (No Dec) and without encoder-decoder structure (No ED).
In No ED, we remove the encoder-decoder from our network and aggregate the features of multi-scale areas in each local region by concatenating (Con) or max pooling (MP).
In Table \ref{table:att}, the result of the attention mechanism with encoder-decoder (Att+ED) is better than the results of removing parts of the attention mechanism and encoder-decoder.
This comparison shows that the decoder works better with the attention mechanism, and the decoder without attention will decrease the capability of the network.
And our RNN-based sequence to sequence structure outperforms hand-crafted manners such as concatenate (Con) and max pooling (MP) in aggregating feature of multi-scale areas.
\begin{table}[htp]
\centering
\caption{The effect of the number of scales $T$ on ModelNet40.}
\label{table:scales}
\begin{tabular}{cccccc}\hline
$T$&4&3&2&1\\ \hline
Acc (\%)&92.54&92.46&\textbf{92.63}&91.94\\ \hline
\end{tabular}
\end{table}

Moreover, we reduce the scale of local regions $T$ from 4 to 1 and remains the largest scale with 128 points.
As depicted in Table \ref{table:scales}, we obtain a even better result of $92.62\%$ with $T=2$ $(K_1=64,K_2=128)$. 
The results with $T > 1$ are better than the result with $T=1$, which shows that the strategy of multi-scale areas can be better in capturing contextual information from local regions.
Therefore, the number of areas $T$ affects the performance of Point2Sequence in extracting the information from local regions.

\begin{table}[!htp]
\centering
\caption{The effect of learning rate on ModelNet40.}
\label{table:lr}
\begin{tabular}{ccccc}\hline
LR&0.0005&0.001&0.002\\ \hline
Acc (\%)&91.94&\textbf{92.63}&91.86\\ \hline
\end{tabular}
\end{table}

Finally, based on the setting of multi-scale areas $T=2$, we explore the effect of learning rate (LR) by setting it to 0.0005 and 0.002.
As shown in Table \ref{table:lr}, the highest accuracy is reached at LR = 0.001.
Therefore, LR = 0.001 is the optimal choice of Point2Sequence.

\subsection{Shape Classification}
In this subsection, we evaluate the performance of Point2Sequence on ModelNet10 and ModelNet40 benchmarks, where
ModelNet10 contains 4899 CAD models which is split into 3991 for training and 908 for testing.
Table \ref{table:compare} compares Point2Sequence with the state-of-the-art methods in the shape classification task on ModelNet10 and ModelNet40.
We compare our method with the results of eight recently ranked methods on each benchmark in terms of class average accuracy and instance average accuracy.
For fair comparison, all the results in Table \ref{table:compare}  are obtained under the same condition, which handles with raw point sets without the normal vector.
By optimizing the cross entropy loss function in the training process, on both benchmarks, Point2Sequence outperforms other methods in class average accuracies and instance average accuracies.
In ModelNet40 shape classification, our method achieves the instance accuracy of 92.6\% which is 1.9\% and 0.2\% higher than PointNet++ (\citealt{qi2017pointnet++}) and DGCNN (\citealt{wang2018dynamic}), respectively.
Experimental results show that Point2Sequence outperforms other methods by extracting the contextual information of local regions.

\subsection{Part Segmentation}
To further validate that our approach is qualified for point cloud analysis, we evaluate Point2Sequence on the semantic part segmentation task.
The goal of part segmentation is to predict semantic part label for each point of the input point cloud.
As dipected in Figure \ref{fig:network_arichitecture}, we build the part segmentation branch to implement the per-point classification.

In part segmentation, ShapeNet part dataset is used as our benchmark for the part segmentation task, the dataset contains 16881 models from 16 categories and is spit into train set, validation set and test set following PointNet++.
There are 2048 points sampled from each 3D shape, where each point in a point cloud object belongs to a certain one of 50 part classes and each point cloud contains 2 to 5 parts.

We employ the mean Intersection over Union (IoU) proposed in (\citealt{qi2017pointnet}) as the evaluation metric. 
For each shape, the IoU is computed between groundtruth and the prediction for each part type in the shape category.
To calculate the mIoU for each shape category, we compute the average of all shape mIoUs in the shape category.
Overall mIoU is also calculated as the average mIoUs over all test shapes.
Similar to the shape classification task, we optimized the cross entropy loss in the training process. 
We compare our results with PointNet (\citealt{qi2017pointnet}), PointNet++ (\citealt{qi2017pointnet++}), Kd-Net (\citealt{klokov2017escape}), SO-Net (\citealt{li2018so}), KC-Net (\citealt{shen2018mining}), ShapeContextNet (\citealt{xie2018attentional}) and DGCNN (\citealt{wang2018dynamic}).
In Table \ref{table:part_segmentaion}, we report the performance of Point2Sequence in each category and the mean IoU of all testing shapes.
Compared with the stated-of-the-art methods, Point2Sequence acquires the best mean instance IoU of $85.2\%$ and achieves the comparable performances on many categories.
Figure \ref{fig:part_segmentation} visualizes some examples of our predicted results, where our results are highly consistent with the ground truth. 
\section{Conclusions}
In this paper, we propose a novel representation learning framework for point cloud processing in the shape classification and part segmentation tasks.
An attention-based sequence to sequence model is proposed to utilize a sequence of multi-scale areas, which focuses on learning the correlation of different areas in a local region. 
To enhance the performance, an attention mechanism is adopted to highlight the importance of multi-scale areas in the local region.
Experimental results show that our method achieves competitive performances with the state-of-the-art methods.

\section{Acknowledgments}
Yu-Shen Liu is the corresponding author. This work was supported by National Key R\&D Program of China (2018YFB0505400), the National Natural Science Foundation of China (61472202), and Swiss National Science Foundation grant (169151). We thank all anonymous reviewers for their constructive comments.

\bibliographystyle{aaai}
\bibliography{reference}

\begin{thebibliography}{}

\bibitem[\protect\citeauthoryear{Bahdanau, Cho, and
  Bengio}{2014}]{bahdanau2014neural}
Bahdanau, D.; Cho, K.; and Bengio, Y.
\newblock 2014.
\newblock Neural machine translation by jointly learning to align and
  translate.
\newblock {\em arXiv:1409.0473}.

\bibitem[\protect\citeauthoryear{Bertolami \bgroup et al\mbox.\egroup
  }{2009}]{bertolami2009novel}
Bertolami, R.; Bunke, H.; Fernandez, S.; Graves, A.; Liwicki, M.; and
  Schmidhuber, J.
\newblock 2009.
\newblock A novel connectionist system for improved unconstrained handwriting
  recognition.
\newblock {\em TPAMI}.

\bibitem[\protect\citeauthoryear{Cho \bgroup et al\mbox.\egroup
  }{2014}]{cho2014learning}
Cho, K.; Van~Merri{\"e}nboer, B.; Gulcehre, C.; Bahdanau, D.; Bougares, F.;
  Schwenk, H.; and Bengio, Y.
\newblock 2014.
\newblock {Learning phrase representations using RNN encoder-decoder for
  statistical machine translation}.
\newblock {\em arXiv:1406.1078}.

\bibitem[\protect\citeauthoryear{Elman}{1990}]{elman1990finding}
Elman, J.~L.
\newblock 1990.
\newblock Finding structure in time.
\newblock {\em Cognitive Science} 14(2):179--211.

\bibitem[\protect\citeauthoryear{Golovinskiy, Kim, and
  Funkhouser}{2009}]{golovinskiy2009shape}
Golovinskiy, A.; Kim, V.~G.; and Funkhouser, T.
\newblock 2009.
\newblock {Shape-based recognition of 3D point clouds in urban environments}.
\newblock In {\em ICCV},  2154--2161.

\bibitem[\protect\citeauthoryear{Hochreiter and
  Schmidhuber}{1997}]{hochreiter1997long}
Hochreiter, S., and Schmidhuber, J.
\newblock 1997.
\newblock Long short-term memory.
\newblock {\em Neural Computation} 9(8):1735--1780.

\bibitem[\protect\citeauthoryear{Kanezaki, Matsushita, and
  Nishida}{2016}]{kanezaki2016rotationnet}
Kanezaki, A.; Matsushita, Y.; and Nishida, Y.
\newblock 2016.
\newblock {RotationNet: Joint object categorization and pose estimation using
  multiviews from unsupervised viewpoints}.
\newblock {\em arXiv:1603.06208}.

\bibitem[\protect\citeauthoryear{Klokov and Lempitsky}{2017}]{klokov2017escape}
Klokov, R., and Lempitsky, V.
\newblock 2017.
\newblock {Escape from cells: Deep kd-networks for the recognition of 3D point
  cloud models}.
\newblock In {\em ICCV},  863--872.

\bibitem[\protect\citeauthoryear{Li and Wu}{2015}]{li2015constructing}
Li, X., and Wu, X.
\newblock 2015.
\newblock Constructing long short-term memory based deep recurrent neural
  networks for large vocabulary speech recognition.
\newblock In {\em ICASSP},  4520--4524.

\bibitem[\protect\citeauthoryear{Li \bgroup et al\mbox.\egroup
  }{2016}]{li2016fpnn}
Li, Y.; Pirk, S.; Su, H.; Qi, C.~R.; and Guibas, L.~J.
\newblock 2016.
\newblock {FPNN: Field probing neural networks for 3D data}.
\newblock In {\em NIPS},  307--315.

\bibitem[\protect\citeauthoryear{Li \bgroup et al\mbox.\egroup
  }{2018}]{li2018pointcnn}
Li, Y.; Bu, R.; Sun, M.; and Chen, B.
\newblock 2018.
\newblock {PointCNN}.
\newblock {\em arXiv:1801.07791}.

\bibitem[\protect\citeauthoryear{Li, Chen, and Lee}{2018}]{li2018so}
Li, J.; Chen, B.~M.; and Lee, G.~H.
\newblock 2018.
\newblock {SO-Net: Self-organizing network for point cloud analysis}.
\newblock In {\em CVPR},  9397--9406.

\bibitem[\protect\citeauthoryear{Luong, Pham, and
  Manning}{2015}]{luong2015effective}
Luong, M.-T.; Pham, H.; and Manning, C.~D.
\newblock 2015.
\newblock Effective approaches to attention-based neural machine translation.
\newblock {\em arXiv:1508.04025}.

\bibitem[\protect\citeauthoryear{Maturana and
  Scherer}{2015}]{maturana2015voxnet}
Maturana, D., and Scherer, S.
\newblock 2015.
\newblock {Voxnet: A 3D convolutional neural network for real-time object
  recognition}.
\newblock In {\em IROS},  922--928.

\bibitem[\protect\citeauthoryear{Qi \bgroup et al\mbox.\egroup
  }{2017a}]{qi2017frustum}
Qi, C.~R.; Liu, W.; Wu, C.; Su, H.; and Guibas, L.~J.
\newblock 2017a.
\newblock {Frustum pointnets for 3D object detection from RGB-D data}.
\newblock {\em arXiv:1711.08488}.

\bibitem[\protect\citeauthoryear{Qi \bgroup et al\mbox.\egroup
  }{2017b}]{qi2017pointnet}
Qi, C.~R.; Su, H.; Mo, K.; and Guibas, L.~J.
\newblock 2017b.
\newblock {PointNet: Deep learning on point sets for 3D classification and
  segmentation}.
\newblock In {\em CVPR}.

\bibitem[\protect\citeauthoryear{Qi \bgroup et al\mbox.\egroup
  }{2017c}]{qi2017pointnet++}
Qi, C.~R.; Yi, L.; Su, H.; and Guibas, L.~J.
\newblock 2017c.
\newblock {PointNet++: Deep hierarchical feature learning on point sets in a
  metric space}.
\newblock In {\em NIPS},  5099--5108.

\bibitem[\protect\citeauthoryear{Riegler, Ulusoy, and
  Geiger}{2017}]{riegler2017octnet}
Riegler, G.; Ulusoy, A.~O.; and Geiger, A.
\newblock 2017.
\newblock {OctNet: Learning deep 3D representations at high resolutions.}
\newblock In {\em CVPR}.

\bibitem[\protect\citeauthoryear{Savva \bgroup et al\mbox.\egroup
  }{2016}]{savva2016shrec}
Savva, M.; Yu, F.; Su, H.; Aono, M.; Chen, B.; Cohen-Or, D.; Deng, W.; Su, H.;
  Bai, S.; Bai, X.; et~al.
\newblock 2016.
\newblock {Shrec’16 track large-scale 3D shape retrieval from ShapeNet
  core55}.
\newblock In {\em Proceedings of the Eurographics Workshop on 3D Object
  Retrieval}.

\bibitem[\protect\citeauthoryear{Shen \bgroup et al\mbox.\egroup
  }{2018}]{shen2018mining}
Shen, Y.; Feng, C.; Yang, Y.; and Tian, D.
\newblock 2018.
\newblock Mining point cloud local structures by kernel correlation and graph
  pooling.
\newblock In {\em CVPR}.

\bibitem[\protect\citeauthoryear{Su \bgroup et al\mbox.\egroup
  }{2015}]{su2015multi}
Su, H.; Maji, S.; Kalogerakis, E.; and Learned-Miller, E.
\newblock 2015.
\newblock {Multi-view convolutional neural networks for 3D shape recognition}.
\newblock In {\em ICCV},  945--953.

\bibitem[\protect\citeauthoryear{Sutskever, Vinyals, and
  Le}{2014}]{sutskever2014sequence}
Sutskever, I.; Vinyals, O.; and Le, Q.~V.
\newblock 2014.
\newblock Sequence to sequence learning with neural networks.
\newblock In {\em NIPS},  3104--3112.

\bibitem[\protect\citeauthoryear{Wang and Posner}{2015}]{wang2015voting}
Wang, D.~Z., and Posner, I.
\newblock 2015.
\newblock Voting for voting in online point cloud object detection.
\newblock In {\em Robotics: Science and Systems}.

\bibitem[\protect\citeauthoryear{Wang \bgroup et al\mbox.\egroup
  }{2018}]{wang2018dynamic}
Wang, Y.; Sun, Y.; Liu, Z.; Sarma, S.~E.; Bronstein, M.~M.; and Solomon, J.~M.
\newblock 2018.
\newblock {Dynamic graph CNN for learning on point clouds}.
\newblock {\em arXiv:1801.07829}.

\bibitem[\protect\citeauthoryear{Wang, Pelillo, and
  Siddiqi}{2017}]{wang2017dominant}
Wang, C.; Pelillo, M.; and Siddiqi, K.
\newblock 2017.
\newblock {Dominant set clustering and pooling for multi-view 3D object
  recognition}.
\newblock In {\em BMVC}.

\bibitem[\protect\citeauthoryear{Wu \bgroup et al\mbox.\egroup
  }{2015}]{wu20153d}
Wu, Z.; Song, S.; Khosla, A.; Yu, F.; Zhang, L.; Tang, X.; and Xiao, J.
\newblock 2015.
\newblock {3D ShapeNets: A deep representation for volumetric shapes}.
\newblock In {\em CVPR},  1912--1920.

\bibitem[\protect\citeauthoryear{Xie \bgroup et al\mbox.\egroup
  }{2018}]{xie2018attentional}
Xie, S.; Liu, S.; Chen, Z.; and Tu, Z.
\newblock 2018.
\newblock {Attentional ShapeContextNet for point cloud recognition}.
\newblock In {\em CVPR},  4606--4615.

\bibitem[\protect\citeauthoryear{Xu \bgroup et al\mbox.\egroup
  }{2018}]{xu2018spidercnn}
Xu, Y.; Fan, T.; Xu, M.; Zeng, L.; and Qiao, Y.
\newblock 2018.
\newblock {SpiderCNN: Deep Learning on Point Sets with Parameterized
  Convolutional Filters}.
\newblock {\em arXiv:1803.11527}.

\bibitem[\protect\citeauthoryear{Zhu \bgroup et al\mbox.\egroup
  }{2017}]{zhu2017target}
Zhu, Y.; Mottaghi, R.; Kolve, E.; Lim, J.~J.; Gupta, A.; Fei-Fei, L.; and
  Farhadi, A.
\newblock 2017.
\newblock Target-driven visual navigation in indoor scenes using deep
  reinforcement learning.
\newblock In {\em ICRA},  3357--3364.

\end{thebibliography}
\end{document}